\pdfoutput=1

\documentclass[11pt]{article}

\usepackage{acl}

\usepackage{times}
\usepackage{latexsym}

\usepackage[T1]{fontenc}

\usepackage[utf8]{inputenc}

\usepackage{microtype}

\usepackage{inconsolata}

\usepackage{tabularx}
\usepackage{graphicx}
\usepackage{lipsum}
\usepackage{url}
\usepackage{tikz}

\title{Stacking the Odds: Transformer-Based Ensemble for AI-Generated Text Detection}

\author{\begin{tabular}{ccccc}
Duke Nguyen & & Khaing Myat Noe Naing & &Aditya Joshi \\
\end{tabular}\\
\begin{tabular}{ccc}
\multicolumn{3}{c}{University of New South Wales, Sydney, Australia}\\
\multicolumn{3}{c}{\tt \{duke.nguyen, khaingmyatnoenaing\}@student.unsw.edu.au, aditya.joshi@unsw.edu.au}
\end{tabular}
}

\begin{document}
\maketitle
\begin{abstract}

This paper reports our submission under the team name `SynthDetectives' to the ALTA 2023 Shared Task. We use a stacking ensemble of Transformers for the task of AI-generated text detection. Our approach is novel in terms of its choice of models in that we use accessible and lightweight models in the ensemble. We show that ensembling the models results in an improved accuracy in comparison with using them individually. Our approach achieves an accuracy score of 0.9555 on the official test data provided by the shared task organisers.

\end{abstract}

\section{Introduction}
Transformer \cite{vaswani2017attention} is a sequence-to-sequence model that has enabled the training of large language models (LLMs). LLMs such as GPT enable text generation in response to user-defined prompts, allowing for wide applicability. As a result, they have proliferated into several aspects of society, both for good and for bad. Text generated from LLMs, when used unethically, can have several detrimental implications: they can cause widespread fake news, dispense away with all notions of academic honesty and authorship, and threaten to replace human-generated information with AI-generated data at large.

Motivated by these existential concerns, many models have been developed to distinguish AI-generated content from human's. ALTA is participatory in tackling this issue by announcing the ALTA 2023 Shared Task, whose goal is to build `automatic detection systems that can discriminate between human-authored and synthetic text generated by Large Language Models (LLMs)' \cite{molla2013overview}. The text comes from a variety of sources in terms of domains (e.g. medical, law), and source model (e.g. GPT-X, T5). Technically, participating teams are required to build an automated system to solve a binary classification task, distinguishing between human and AI-generated text. Models are evaluated based on robustness and accuracy. There is no requirement on the efficiency and run-time performance. Our team participated in the said shared task. The code is available here\footnote{\url{https://github.com/dukeraphaelng/synth_detectives}}. We stack multiple Transformer-based models in an ensemble and show that the ensemble performs better than the individual models. In this paper, we will discuss existing works in the domain, our analysis of the original training data, our proposed pipeline and architecture, our experimental results, and suggested future work.

\section{Related Work} \label{sec:related_works}
AI-generated text detection has a long history. The sources of our AI-generated text are LLMs, which constrain our task to `authorship attribution (AA) for neural texts', also known as Neural Text Detection (NTD). It is a subclass of the task of binary classification (and sometimes multi-class, when we are detecting the source model). We will summarise briefly the current literature in this domain. Our main source comes from two major surveys by ~\citet{jawahar2020} and ~\citet{uchendu2023}. The latter classifies automated NTD as follows:

\textbf{Stylometric attribution} detects Neural Text Generator (NTG) using ensembles of classical machine learning (ML) models trained on stylometric features such as LIWC (Linguistic Inquiry \& Word Count), POS tags, n-grams, Readability score, WritePrints, Empath. These models work best on a small dataset. However, as we increase the data size, they are outperformed by deep learning models rapidly.

\textbf{Deep learning: GLoVe-based attribution}: GLoVe ~\cite{glove} is an unsupervised learning algorithm that aggregates global word-word co-occurrence statistics from text to build word representation. GLoVe-based models use these embeddings with RNN and LSTM, which was considered SOTA before BERT~\cite{bert}.

\textbf{Deep learning: Energy-based attribution}: Energy-based models (EBMs)~\cite{energy} are `un-normalized generative models' using some energy function to generate high-quality data by modelling the probability distribution of the training data. Adapted for NTD~\cite{energydetect}, they perform well on unseen data, however, they do not scale as well, and are very expensive to train.

\textbf{Deep learning: Transformer-based attribution}: is Transformer-based models fine tuned to perform NTD. These models surpass stylometric and GLoVe-based models and are cheaper than EBMs. RoBERTa and BERT are two models that frequently achieve high performance on NTD benchmarks \cite{uchendu2023}. Other Transformer-based models that are used in NTD include ELECTRA, XLNet, and DeBERTa. These inspire our choice of weak learners.

\textbf{Statistical attribution}: was developed to combat top-p and top-k decoding strategies which Transformers are not well-equipped against. It has been shown that `human language is stationary and ergodic as opposed to neural language' \cite{varshney2020} suggesting the validity of this approach. Four different algorithms have been proposed which detect AI-generated text through statistical distributions. These are: GLTR \cite{gltr}, MAUVE \cite{mauve}, Distribution detector \cite{distdetect}, and DetectGPT \cite{detectgpt}, the last three of which perform competitively.

\textbf{Hybrid attribution}: is ensembles using several previously described detectors. These include TDA-based detector~\cite{tda}, which extracts attention matrices of BERT's word representations and process them through TDA-based methods as features for a logistic regression model, Fingerprint detector~\cite{fingerprint}, which ensembles fine-tuned RoBERTa embeddings and CNN classifier), FAST~\cite{fast}, which uses RoBERTa with a Graph Neural Network), and CoCo~\cite{coco}, a coherence-based contrastive learning model. Our work is an ensemble-based approach to the task. However, we use an ensemble of Transformer models.

\section{Dataset}
Three subsets of the dataset are presented: training, validation, and testing. The training set contains 18,000 entries, and the validation and testing each contains 2,000 entries. Evaluation is based on the testing set which was not released until the testing phase of the competition. The training set contains three columns `id', `text', and `label' (1 if human-generated, 0 if AI-generated). The validation and testing set each contains two columns `id', and `text'. Samples of the training set are shown in Table \ref{tab:sample_train}.

\begin{table}
    \centering
    \begin{tabularx}{0.5\textwidth}{|c|X|c|}
        \hline
        id & text & label \\
        \hline
        0 & `Have you ever heard of the Crusades? A time in which Christians went on a 200 year rampage throughout Europe and on their path to Isreal in which they slaughtered innocent people in the name of your God?' & 1 \\
        \hline
        4 & `The Circuit Court of Appeals of New Jersey had jurisdiction of the controversy between these parties, and its decree was affirmed. But as the court had jurisdiction under the original act of Congress, the jurisdiction in this case was also, under the act of Congress, a bar to the suit.' & 0 \\
        \hline
    \end{tabularx}
    \caption{Samples from the training set.}
    \label{tab:sample_train}
\end{table}

When analyzing the dataset, we find that the AI-generated and human-generated text is evenly split into 9000-9000 entries respectively in the training set. We also find that the average word count per text is relatively low. The mean length is in the 34-35 range in the three subsets, with a standard deviation in the 26.7-27.9 range, a maximum of 172-193, and a minimum of 1, making this a short sequence task.

To find the main domains of the text, we remove all stop words from each set and find the frequency of n-gram phrases from the cleaned corpus, and pick the top-k elements from each set. We look at the n-gram range of $(3,4)$, with $k=10$. We find that overwhelmingly all the phrases are in the domain of law across the three sets. The following list is the union of the three sets with the above configuration: \{`court of appeals', `of the court', `of the united', `of the united states', `opinion of the', `the court of', `the court of appeals', `the district court', `the opinion of', `the united states'\}.
\section{Approach}
\begin{figure}
    \centering
    \begin{tikzpicture}[node distance={20mm}, thick, main/.style = {draw}] 
    \node[main] (i) {Text};
    \node[main] (3) [below right of=i] {RoBERTa};
    \node[main] (2) [left of=3] {ELECTRA};
    \node[main] (1) [left of=2] {ALBERT};
    \node[main] (4) [right of=3] {XLNet};

    \node[main] (m) [below left of=3] {Logistic Regression};
    \node[main] (o) [below of=m] {Prediction};

    \draw[->] (i) -- (1);
    \draw[->] (i) -- (2);
    \draw[->] (i) -- (3);
    \draw[->] (i) -- (4);

    \draw[->] (1) -- (m);
    \draw[->] (2) -- (m);
    \draw[->] (3) -- (m);
    \draw[->] (4) -- (m);

    \draw[->] (m) -- (o);

    \end{tikzpicture} 
    \caption{Our Stack Ensembling Architecture for AI-generated text detection.}
    \label{fig:architecture}
\end{figure}
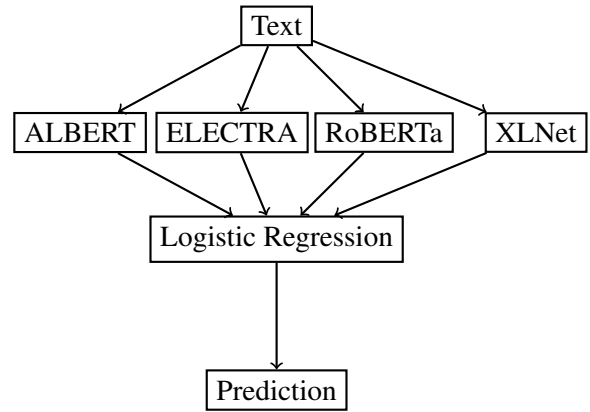

Our approach uses a stacking ensemble of classifiers (as shown in Figure \ref{fig:architecture}) to perform our training, validation, and testing. A stacking ensemble of classifiers acts similarly to a weighted voting classifier. Our choice of architecture is inspired by \citet{maloyan22}, which achieved high performance in the RuATD Shared Task 2022 on Artificial Text Detection
in Russian \cite{Shamardina_2022}.

We train each weak classifier using the above dataset split, and then we concatenate the raw predictions on the training set together and feed them to the meta-learner. We use a simple Logistic Regressor as our meta-learner. Our criteria for picking models are ease of use, short-sequence-task-based models, and variety in model architecture. We also choose only encoder-only models, since they are built for regression/ classification tasks, and we can conveniently extract the \texttt{[CLS]} token from their last hidden state to perform Logistic Regression. As a result, we use ALBERT, ELECTRA, RoBERTa and XLNet as the Transformer-based models.

To optimise the training cycle, we tokenise the entire dataset (with the respective model's tokeniser), and pass them through their respective pre-trained model to obtain the \texttt{[CLS]} token from the last hidden state. We consider this to be our dataset and do our splitting, training, and testing on this processed dataset. To train, we pass the \texttt{[CLS]} token through a single fully connected layer, with the input dimension equivalent to the model's \texttt{[CLS]}'s dimension, and the output dimension of 2, then we softmax the output. After fine-tuning the weak models, we perform inference on the training split and concatenate the predictions which are fed for the meta-learner to train.

\section{Experiment Setup}

\subsection{Setup}
We do not perform any data preprocessing on the dataset. We have a train-validation-test split of 0.8, 0.1, 0.1. All training was done on Google Cloud Platform's Vertex Colab GPU for GCE usage on NVIDIA A100 (40 GB).

\subsection{Pipeline}

For both our weak learners and our meta-model, we use the AdamW optimiser with the default settings, i.e. $lr=0.001, \beta=(0.9, 0.999), \epsilon=1e-08, weight\_decay=0.01$. All models are trained with $epochs = 300$ and $batch\_size = 128$. All models in the ensemble are pre-trained models available on HuggingFace (as of 25th October 2023). For each model, we include their architecture name and the unique HuggingFace model identifier associated with their pre-trained weights.

\textbf{ALBERT} (albert-base-v2) \cite{lan2020albert} is a modification of BERT~\cite{bert} which reduces its memory consumption and increases the training speed by repeating layers split among groups and splitting the embedding matrix into smaller matrices, whilst being more performative than BERT in GLUE, RACE, and SQuAD.

\textbf{ELECTRA} (google/electra-small-discriminator) \cite{clark2020electra} is another modification of BERT that changes the pretraining objective, as inspired by GAN where ELECTRA acts as the discriminator which predicts whether a token in a randomly masked text is original or generated by the generator (which we train simultaneously). This approach makes ELECTRA perform comparably to larger models whilst using a lot less compute.    

\textbf{RoBERTa} (roberta-base) \cite{liu2019roberta} optimises BERT in four aspects of training: using full-sentences without Next Sentence Prediction (NSP) loss, with dynamic masking, with larger mini-batches, and with a larger byte-level Byte-Pair Encoding (BPE).

\textbf{XLNet} (xlnet-base-cased) \cite{yang2020xlnet} uses a generalised autoregressive pretraining method that maximises the `expected likelihood over all permutations of the input sequence factorisation order' enabling bidirectional contexts and overcoming BERT's pretrain-finetune discrepancy due to neglecting masked positions dependency. XLNet also builds on Transformer-XL \cite{transformerxl}, outperforming BERT on 20 tasks.

\begin{figure}[ht]
    \centering
    \includegraphics[width=\linewidth]{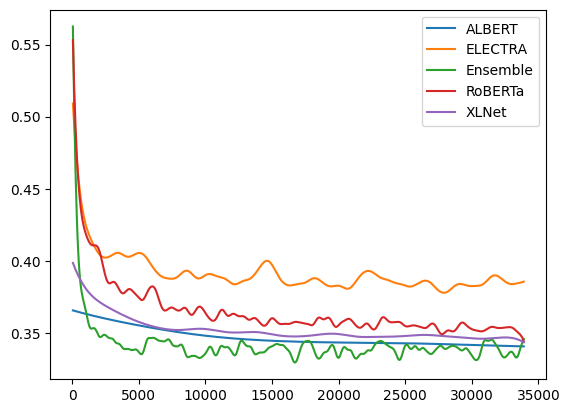}
    \caption{Training Loss.}
    \label{fig:training-loss}
\end{figure}

\begin{figure}[h]
    \centering
    \includegraphics[width=\linewidth]{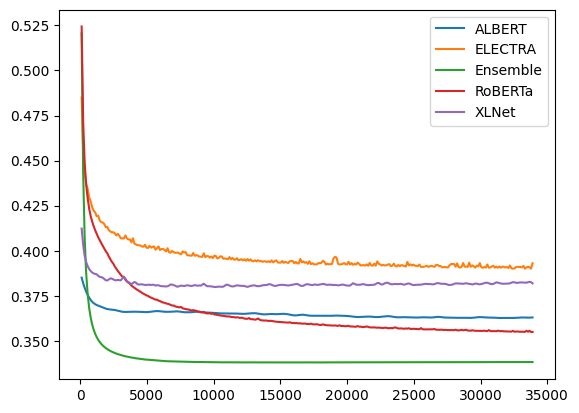}
    \caption{Validation Loss.}
    \label{fig:val-loss}
\end{figure}

\begin{table}[h]
    \centering
    \begin{tabular}{|l|l|}
        \hline
        Model & Accuracy\\
        \hline
        ELECTRA & 0.9311\\
        \hline
        XLNet & 0.9361\\
        \hline
        ALBERT & 0.9567\\
        \hline
        RoBERTa & 0.9572\\
        \hline
        Ensemble & 0.9694\\
        \hline
    \end{tabular}
    \caption{Model Accuracy on the Test Set.}
    \label{tab:accuracy}
\end{table}

\section{Results}
Figure \ref{fig:training-loss} shows our smoothed training loss using the Gaussian kernel (since the original training loss displays too wide short-cycle variation, obfuscating the overall trend), Figure \ref{fig:val-loss} shows our validation loss which follows a similar pattern. Table \ref{tab:accuracy} shows our accuracy on the test set as described in the experiment setup. Among our weak learners, RoBERTa performs the best, followed by ALBERT, XLNet, and finally ELECTRA. As expected, our meta-model (Ensemble) outperforms even RoBERTa by more than $0.012$. The final testing accuracy model ranking is reflected in the validation loss, and to a lesser extent in the training loss. This agrees with much of the literature indicating that RoBERTa is the best learner in AI-generated text prediction \cite{jawahar2020}. We also note that XLNet and ALBERT start with extremely low loss, suggesting their pre-training procedure might be conducive to AI-generated text detection. 

Finally, the ALTA shared task organisers provided us with a shared task test set \textit{i.e.}, the official test set. We achieve an accuracy of $0.9555$ with our stacking ensemble on the official test set.

\section{Conclusion \& Future Work}
In this paper, we describe our system for the ALTA Shared Task 2023. We show how an ensemble of Transformer-based models can be combined using a logistic regression classifier to predict if a text was generated by AI. We achieved an accuracy of $0.9555$ using a stacking ensemble of basic encoder-only Transformer models. 

Our work presents a novel approach to ensemble Transformer-based models to approach the ALTA shared task. However, this work identifies several potential directions for future work. Ensembling models usually benefit from a variety of learners specialised in different types of inputs. We only implemented an ensemble of Transformer classifiers, but it would be beneficial to integrate other non-Transformer-based weak learners as detailed in Section \ref{sec:related_works}. Especially useful would be to integrate contrastive learning in our training procedure. In addition, it would also be useful to perform data augmentation which can help generalise the model. One suggested technique is `text continuation', where given a human-generated text, we slice the first $n$ words and have an LLM finish the sentence. Furthermore, the scope of the shared task does not imply the possibility of an adversarial attack. It has been shown that the RoBERTa detector can be attacked easily through misspelling \cite{wolff2022attacking}. It would also be helpful to build detectors that are resilient in this regard.

\section*{Acknowledgements}

We thank Jake Renzella and the School of Computer Science and Engineering at the University of New South Wales, Sydney for providing us with GPU cloud credits, without which we would not be able to run our experiments. 
\bibliography{ref} 




\end{document}